\documentclass[letterpaper, 10 pt, conference]{ieeeconf}  % Comment this line out if you need a4paper

\IEEEoverridecommandlockouts                              % This command is only needed if 
\usepackage{graphicx} % for pdf, bitmapped graphics files
\usepackage{amssymb}  % assumes amsmath package installed
\usepackage{multirow}
\usepackage{subcaption}

\title{\LARGE \bf
LiDAR-Guided Monocular 3D Object Detection for Long-Range Railway Monitoring
}

\author{Raul David Dominguez Sanchez$^{1,2}$, Xavier Diaz Ortiz$^{1}$, Xingcheng Zhou$^{2}$, Max Peter Ronecker$^{1,3}$, \\ Michael Karner$^{1}$, Daniel Watzenig$^{3}$ and Alois Knoll$^{2}$% <-this % stops a space
\thanks{$^{1}$SETLabs Research GmbH, 80687 Munich, Germany
        {\tt\small \{firstname.lastname\}@setlabs.de}}%
\thanks{$^{2}$Chair of Robotics, Artificial Intelligence and Real-time Systems, Technical University of Munich, 80333 Munich, Germany
        {\tt\small firstname.lastname@tum.de}}%
\thanks{$^{3}$Institute of Visual Computing, Graz University of Technology, 8010 Graz, Austria
        {\tt\small firstname.lastname@tugraz.at}}%
\thanks{Accepted for the Data-Driven Learning for Intelligent Vehicle Applications Workshop at the 36th IEEE Intelligent Vehicles Symposium (IV) 2025}%
}

\begin{document}

\maketitle
\thispagestyle{empty}
\pagestyle{empty}

%%%%%%%%%%%%%%%%%%%%%%%%%%%%%%%%%%%%%%%%%%%%%%%%%%%%%%%%%%%%%%%%%%%%%%%%%%%%%%%%
\begin{abstract}

Railway systems, particularly in Germany, require high levels of automation to address legacy infrastructure challenges and increase train traffic safely. A key component of automation is robust long-range perception, essential for early hazard detection, such as obstacles at level crossings or pedestrians on tracks. Unlike automotive systems with braking distances of ~70 meters, trains require perception ranges exceeding 1 km. This paper presents an deep-learning-based approach for long-range 3D object detection tailored for autonomous trains. The method relies solely on monocular images, inspired by the Faraway-Frustum approach, and incorporates LiDAR data during training to improve depth estimation. The proposed pipeline consists of four key modules: (1) a modified YOLOv9 for 2.5D object detection, (2) a depth estimation network, and (3-4) dedicated short- and long-range 3D detection heads. Evaluations on the OSDaR23 dataset demonstrate the effectiveness of the approach in detecting objects up to 250 meters. Results highlight its potential for railway automation and outline areas for future improvement.

\end{abstract}

%%%%%%%%%%%%%%%%%%%%%%%%%%%%%%%%%%%%%%%%%%%%%%%%%%%%%%%%%%%%%%%%%%%%%%%%%%%%%%%%
\section{INTRODUCTION}
\label{sec:introduction}

Railways are a key mode of transport in Europe, facilitating the movement of goods and passengers efficiently. However, the system faces capacity limits and legacy issues \cite{Borghini2021}. Initiatives such as Digital Schiene Deutschland \cite{digitale_schiene_deutschland}, Shift2Rail \cite{shift2rail}, and safe.trAIn \cite{safetrain} aim to modernize railway operations and increase automation.

A crucial requirement for autonomous rail systems is a robust perception module capable of detecting objects and locating the train with high precision. This is particularly important for improving safety, as most railway accidents involve track intrusions (61\%) or level crossings (24\%) \cite{RailwaySafetyStatistics}. Furthermore, the long braking distances of trains, reaching lengths as high as 1000 meters \cite{ristic-durrantReviewVisionBasedOnBoard2021a,assafHighPrecisionLowCostGimballing2022}, necessitate early detection of obstacles, significantly beyond the typical 70-meter braking distance for cars \cite{yang2024improvingdistant3dobject}.

LiDAR provides accurate depth information but is limited in range (250-500 meters), while monocular cameras capture high-resolution images but lack direct depth measurement \cite{Wang2023}. The fusion of both sensors has shown significant improvements in 3D object detection \cite{wuVirtualSparseConvolution2023,liLoGoNetAccurate3D2023,liVoxelFieldFusion2022,liHomogeneousMultimodalFeature2022}. This work proposes a monocular 3D object detection method tailored for railway environments with the following contributions:

%\begin{itemize}
%    \item Develop an AI-based 3D object detection method for %autonomous railway vehicles using only monocular images.
%    \item Enable long-range detection, up to 250 meters, %addressing railway safety requirements.
%    \item Leverage LiDAR-based 3D labels during training to %improve depth estimation and object localization.
%\end{itemize}

\textbf{Contributions}

\begin{enumerate}
\item An AI-based 3D object detection approach for autonomous railway vehicles using only monocular images, addressing long-range perception challenges.

\item A novel pipeline integrating a modified YOLOv9 for 2.5D detection with depth estimation module for accurate frustum-based 3D localization.

\item A training strategy leveraging LiDAR point clouds and 3D labels to enhance depth estimation and improve monocular-based 3D detection, reducing the gap with LiDAR-based methods.
\end{enumerate}

\section{RELATED WORK}
The following provides a brief introduction to 3D object detection, including relevant detector families such as frustum-based methods and the YOLO family, as well as an overview of depth estimation.

\subsection{3D Object Detection}
\label{02_rel:3d_obj}

3D object detection involves identifying objects in a scene and estimating their enclosing 3D bounding boxes. Typically, 8 to 11 parameters are predicted, including the object class, center coordinates, dimensions, and orientation. In autonomous driving, orientation prediction is often simplified to yaw rotation, as objects are assumed to rotate only about the axis perpendicular to the road \cite{OLEKSIIENKO2022313}.

Methods for 3D object detection are categorized by:
\begin{itemize}
    \item \textbf{Input data type}: Input data typically includes images or point clouds, which may be further processed into formats like depth maps, voxel grids, or BEV maps. Camera-based methods are divided into monocular \cite{yang2024improvingdistant3dobject, zhangMonoDETRDepthguidedTransformer2023}, stereo \cite{youPseudoLiDARAccurateDepth2020}, and multi-camera \cite{liBEVDepthAcquisitionReliable2023} approaches. Multimodal methods fuse data from multiple sensors, such as combining LiDAR and monocular images, to improve performance \cite{wuVirtualSparseConvolution2023,liHomogeneousMultimodalFeature2022}.
    \item \textbf{Detection environment}: Detection methods are tailored to either indoor or outdoor scenes. Indoor point clouds are dense and provide rich context, while outdoor point clouds are sparse and influenced by external factors. Recent work explores environment-agnostic methods \cite{li2024unified3dobjectdetection}.
    \item \textbf{Detection approach}: Deep learning dominates the field, though classical methods remain relevant for efficiency and simpler parameterization. Hybrid methods combine classical techniques (e.g., preprocessing) with deep learning for improved performance \cite{RAULPAPER}.
\end{itemize}

This classification is not standardized, but it highlights the factors most relevant to this work.

\subsection{Frustum-Based Methods for 3D-Detection}
In the context of this paper, a frustum refers to a portion of a 3D point cloud that is pyramid-shaped and is used to narrow down a region of interest in a point cloud.

Frustum-based methods leverage 2D object detections to narrow the 3D search space by projecting 2D bounding boxes into 3D frustums \cite{qiFrustumPointNets3D2018}. Faraway-Frustum \cite{zhangFarawayFrustumDealingLidar2021} combines monocular images and LiDAR point clouds to address the challenges of long-range 3D detection. It uses a dual-head architecture: one head employs standard methods like PV-RCNN \cite{shi2021pvrcnnpointvoxelfeatureset} or F-PointNet \cite{qiFrustumPointNets3D2018} for short-range objects, while the second head (FF-Net) processes long-range frustums in BEV format using MobileNet \cite{howard2017mobilenetsefficientconvolutionalneural}.

FF-Net handles sparsity in long-range LiDAR data and predicts 3D bounding boxes, including yaw orientation, object center, and dimensions. Frustums are classified as short- or long-range based on their centroid's distance, with thresholds (e.g., 60\,m for pedestrians, 75\,m for cars) derived from the KITTI dataset \cite{KITTI}.

By reducing search space and leveraging multimodal data, frustum-based methods improve detection accuracy and efficiency, making them suitable for autonomous driving applications.
\subsection{The YOLO Object Detector Family and Extensions}
\label{subsec:yolo}

The You Only Look Once (YOLO) family of object detectors simplifies 2D detection by combining classification and localization into a single network \cite{7780460}. YOLOv1 predicts bounding boxes and class probabilities on a 7$\times$7 grid, with non-maximal suppression filtering low-confidence predictions. Subsequent versions introduced significant improvements in accuracy, efficiency, and feature extraction.

YOLOv9 \cite{wang2024yolov9learningwantlearn} replaced fixed bounding box dimensions with probabilistic distributions, eliminating non-maximum suppression through one-to-one label assignment. Architectural upgrades include Spatial Pyramid Pooling (SPP), Path Aggregation Networks (PANet), Generalized Efficient Layer Aggregation Networks (GELANs), and Programmable Gradient Information (PGI), enhancing multi-scale feature extraction and gradient flow while maintaining real-time performance \cite{wang2024yolov1yolov10fastestaccurate}.

YOLO has also been adapted for 2.5D tasks, integrating depth estimation. YOLO-D and Dist-YOLO extend YOLOv3 to directly predict depth in the detection head \cite{haseebMachineLearningTechniques2021,app12031354}. Another approach overlays YOLOv5 bounding boxes on depth maps to estimate distances via polynomial regression \cite{Masoumian_2021}. While effective for short ranges, these methods require recalibration for new domains.

\subsection{Depth Estimation}
\label{subsec:depth_estimation}

Monocular depth estimation converts 2D images into depth maps, enabling point cloud generation. Classical methods relied on geometric cues such as structure-from-motion and multi-view stereo, but deep learning now dominates due to its ability to infer depth from single images. Supervised approaches use ground truth depth, while self-supervised methods exploit geometric consistency. The following methods are relevant to this work. MiDAS \cite{ranftl2020robustmonoculardepthestimation} trains on multiple datasets for cross-domain generalization using zero-shot transfer. Fine-tuning is required for absolute depth estimation. DenseDepth \cite{Alhashim2018} uses an encoder-decoder with a DenseNet \cite{huang2018denselyconnectedconvolutionalnetworks} backbone. Its loss function combines L1, gradient, and SSIM terms. Sparse ground truth is addressed with "Colorization using Optimization" \cite{levin2004colorization}. Marigold \cite{ke2023repurposing} applies generative diffusion models, treating depth estimation as a denoising process. It improves generalization but increases computational cost.

\subsection{Railway Datasets}
\label{subsec:railway_datasets}

Public datasets for autonomous railway perception are limited compared to the automotive domain, where large-scale datasets like KITTI \cite{KITTI}, NuScenes \cite{nuscenes}, and Waymo Open Dataset \cite{WOD} dominate 3D object detection research. Existing railway datasets vary in focus, covering tasks such as 2D semantic segmentation \cite{zendelRailSem19DatasetSemantic2019, 8859360}, object detection \cite{harb2020frsignlargescaletrafficlight, 9715050, 10052883}, and 3D perception \cite{9852821, khemmarRoadRailwaySmart2022, Neri_2022_EUVIP}. However, many are either not publicly available \cite{haseebMachineLearningTechniques2021, khemmarRoadRailwaySmart2022}, lack mainline railway scenarios, or do not provide long-range annotations. 

For long-range 3D object detection, datasets with labeled objects beyond 200 meters are necessary. While some datasets, such as ESRORAD \cite{khemmarRoadRailwaySmart2022} and LRODD \cite{haseebMachineLearningTechniques2021}, include multimodal data, they either prioritize road environments or remain inaccessible. 

OSDaR23 \cite{osdar23} addresses the lack of public railway datasets by providing high-quality sensor data for 2D and 3D object detection. It includes six RGB cameras (high/low resolution), three infrared cameras, six LiDAR sensors, a 2D radar, and GNSS/IMU.

The dataset comprises 1534 annotated frames across 21 sequences and 45 subsequences. Annotations, following the ASAM OpenLABEL standard \cite{Openlabel}, include 2D bounding boxes, polylines for track recognition, 3D cuboids, and per-point semantic segmentation. It is divided into training (~70\%), validation (~15\%), and test (~15\%) sets. Given all of this, OSDaR23 provides the necessary data for our research objectives in long-range 3D object detection railway environments.

\begin{figure*}[thpb]
     \centering
     \begin{subfigure}[b]{\textwidth}
         \centering
         \includegraphics[width=\textwidth]{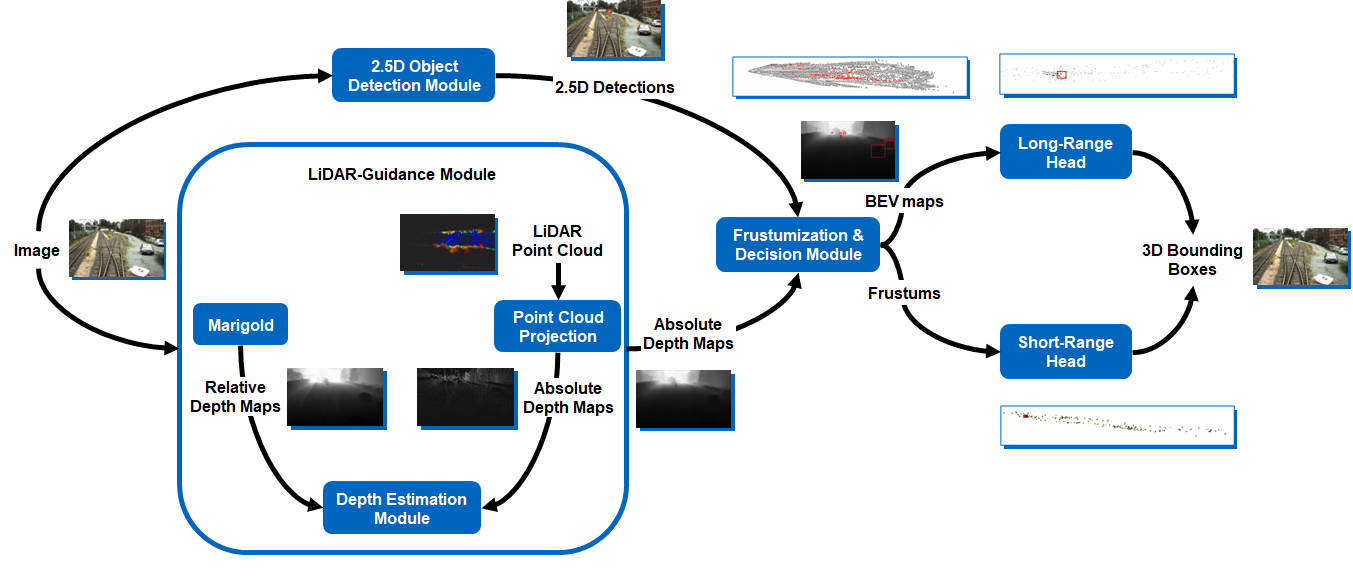}
         %\caption{}
     \end{subfigure}
     \begin{subfigure}[b]{\textwidth}
         \centering
         \includegraphics[width=\textwidth]{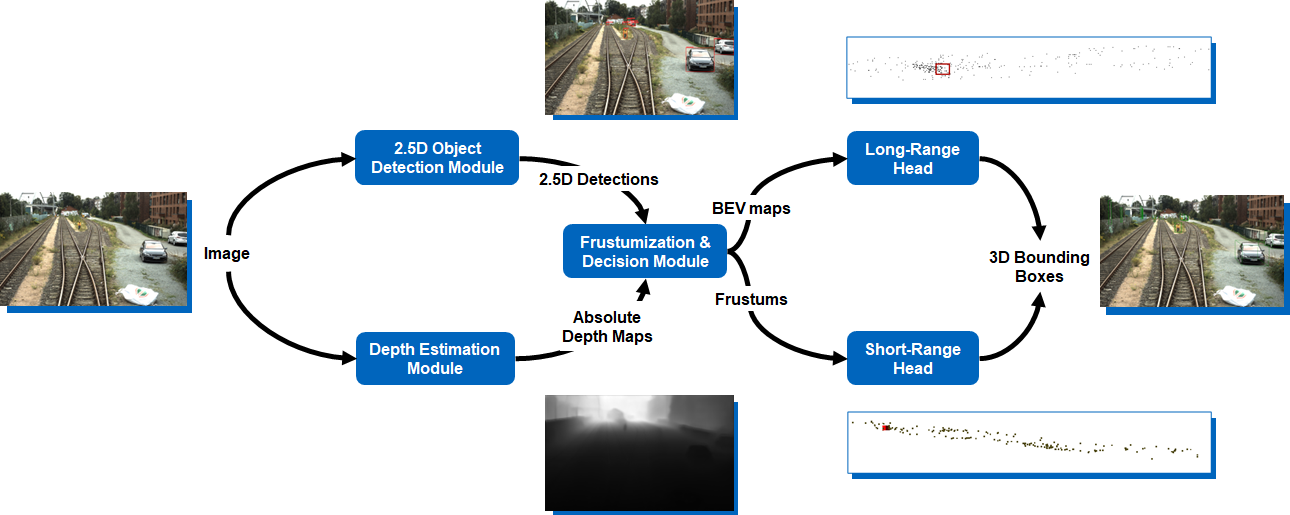}
         %\caption{}
     \end{subfigure}
     \caption{MFF Pipeline. Top: Pipeline during training, when LiDAR point clouds are used to guide depth estimation. Bottom: Pipeline during testing time, in this case the method works solely using monocular images.}
    \label{fig:Pipeline}
\end{figure*}

\section{METHODOLOGY}
Given the objectives established for this project, as well as the reviewed literature, the Monocular Faraway-Frustum pipeline (MFF) for 3D object detection was devised (shown in Fig. \ref{fig:Pipeline}), which leverages LiDAR data during training to enhance object depth estimation, especially for distant objects (in this work distant objects are those beyond 100 meters). In the following subsections the different modules of the developed pipeline will be described.

\subsection{Depth Estimation Module}
As can be inferred from the name, the main inspiration for MFF is Faraway-Frustum mainly due to its emphasis on improving long-range detection. Nevertheless, since Faraway-Frustum is a multi-modal approach (monocular images and LiDAR point clouds) and the proposed method is monocular-based, it was decided to replace the LiDAR point cloud input with pseudo-clouds generated by back-projecting depth maps. Therefore, a monocular depth estimation module was necessary to estimate depth maps. In monocular depth estimation research, diffusion-based approaches (e.g. Marigold) are currently the state-of-the-art. However, their main disadvantage is their inability of running in real-time. Furthermore, works like MiDAS emphasize the utilization of diverse and large sets of data for training as well as to initially train for relative depth estimation and then fine-tune for absolute depth estimation. As previously mentioned, the mainly utilized dataset in this work was OSDaR23, which does not provide ground truths for relative or absolute depth estimation (the latter can be computed from point clouds). Consequently, to take advantage of the SOTA for depth estimation, to be capable of running the module in real-time, and follow the "relative then absolute" approach, first, given monocular images, from OSDaR23 and RailSem19 \cite{zendelRailSem19DatasetSemantic2019} (cropped to have the same ratio as OSDaR23), Marigold is used as the teacher network to guide the outputs of a modified version of DenseDepth. The modifications include a change in the depth loss from L1 to Huber, and a switch in the upsampling method from bilinear interpolation to transposed convolutions. After successful distillation the second step is to train a simple convolutional network that was designated as refinement network. What this network does is map the relative depth to absolute depth. Additionally, in this case the ground truths were inpainted depth maps (using the optimization method described in \cite{levin2004colorization}) generated from OSDaR23's point clouds.

\subsection{2.5D Object Detection Module}
In Faraway-Frustum a 2D object detector is used to identify regions of interest in point clouds (frustums). Based on the centroid of a frustum, it is decided whether this cloud should be processed with a short-range detection head or a long-range detection head. Even though the same steps could be followed in MFF, due to the fact that pseudo-clouds are not as accurate as LiDAR point clouds and that they tend to include more noise, it was decided to do a weighted sum instead. The summed components are a frustum's centroid and an estimation of the distance of the object in such frustum. To do so, our method replaces the 2D object detector with a 2.5D object detector, that is a 2D object detector that also estimates the distance to the camera for the predicted objects. Works like \cite{haseebMachineLearningTechniques2021,app12031354} modified YOLO to also predict distance, based on such works the following modifications were applied to YOLOv9 (the latest version of the YOLO detectors at the time) to create a 2.5D object detection module:
\begin{itemize}
    \item \textbf{Distance Head}: Besides the box regression head and the class head an additional head was added comprised of a simple fully convolutional network and producing normalized values which would represent distances between 0 and 250 meters.
    \item \textbf{Huber Loss}: The usually used losses for distance estimation are either an L1 loss or L2 loss. When used individually, the losses did not produce estimations as accurate as expected and tended to overfit to the distance range with the most examples. It was decided to combine them using Huber Loss which resulted in a better performance. 
    \item \textbf{Multiple Datasets}: Since OSDaR23 is a relatively small dataset, it was difficult to reach a good performance using solely this dataset. In works like \cite{haseebMachineLearningTechniques2021}, they integrate data from KITTI to boost the performance of the 2.5D object detector. Therefore, a similar approach was followed by training this module with OSDaR23 and KITTI. Additionally, KITTI+OSDaR23 training was used as a pre-training step, after which the model was fine-tuned only using OSDaR23. 
\end{itemize}

\subsection{Frustumization \& Decision Module}
The next block in the pipeline is the frustumization and decision block, which takes as input a depth map and 2.5D object predictions, based on these it generates frustums and estimates the center of the frustums (using the aforementioned weighted sum), to then transform these clouds from the LiDAR frame to the frustum frame and feed them to the corresponding head. In the case of the long-range head, the frustums are splatted into Bird's Eye View (BEV) representation before being fed to the head together with their respective class priors. 
\subsection{Detection Heads}
Similarly to Faraway-Frustum, the long-range head in MFF is comprised of a feature encoder and fully connected layers. However, in contrast with the original implementation, MFF utilized various sets of fully connected layers to increase the capacity of the module to represent more complex functions and better detect the classes of interest. On the other hand, for the short-range head and with the objective of leveraging SOTA point-cloud-based 3D object detectors (since they are usually developed to work in a short-range detection environment), Faraway-Frustum utilizes PV-RCNN, with that in mind, it was decided to evaluate the performance of the pipeline with different heads, being these: PointRCNN, PointPillars, and Part-A$^2$. Both modules were only trained with the OSDaR23 dataset.

\section{Evaluation}
Since each module was trained separately, we first evaluate their individual performance before assessing their integration into the full pipeline. Individual evaluations use the OSDaR23 validation set, while full pipeline evaluations use its test split, ensuring design choices are guided by performance gains while maintaining independent assessment. 

\begin{figure}[thpb]
      \centering
      \parbox{3.3in}{
          \centering
          \begin{subfigure}[b]{0.485\textwidth}
             \centering
             \includegraphics[width=\textwidth]{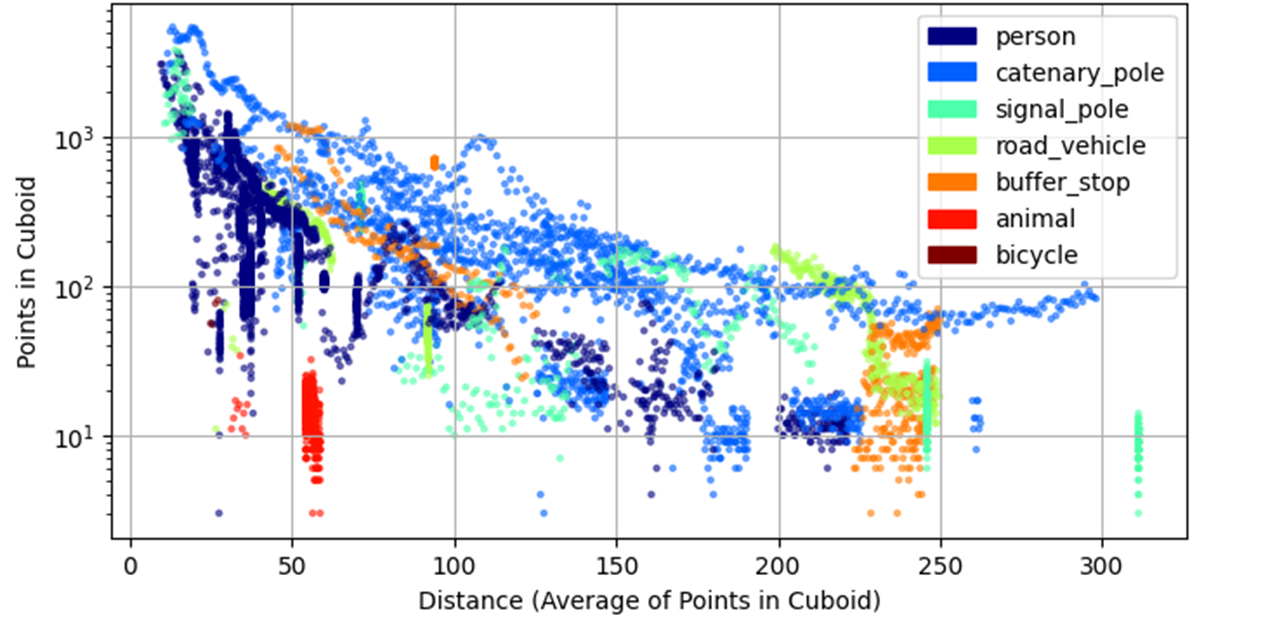}
          \end{subfigure}
          \begin{subfigure}[b]{0.485\textwidth}
             \centering
             \includegraphics[width=\textwidth]{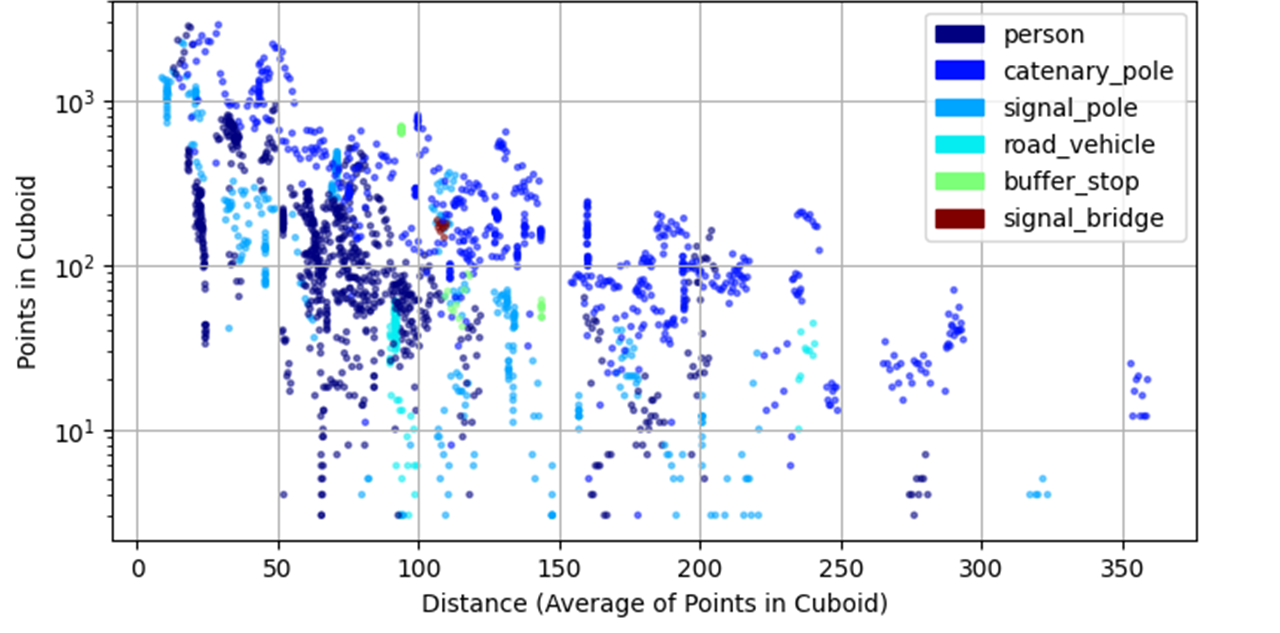}
          \end{subfigure}
          \begin{subfigure}[b]{0.485\textwidth}
             \centering
             \includegraphics[width=\textwidth]{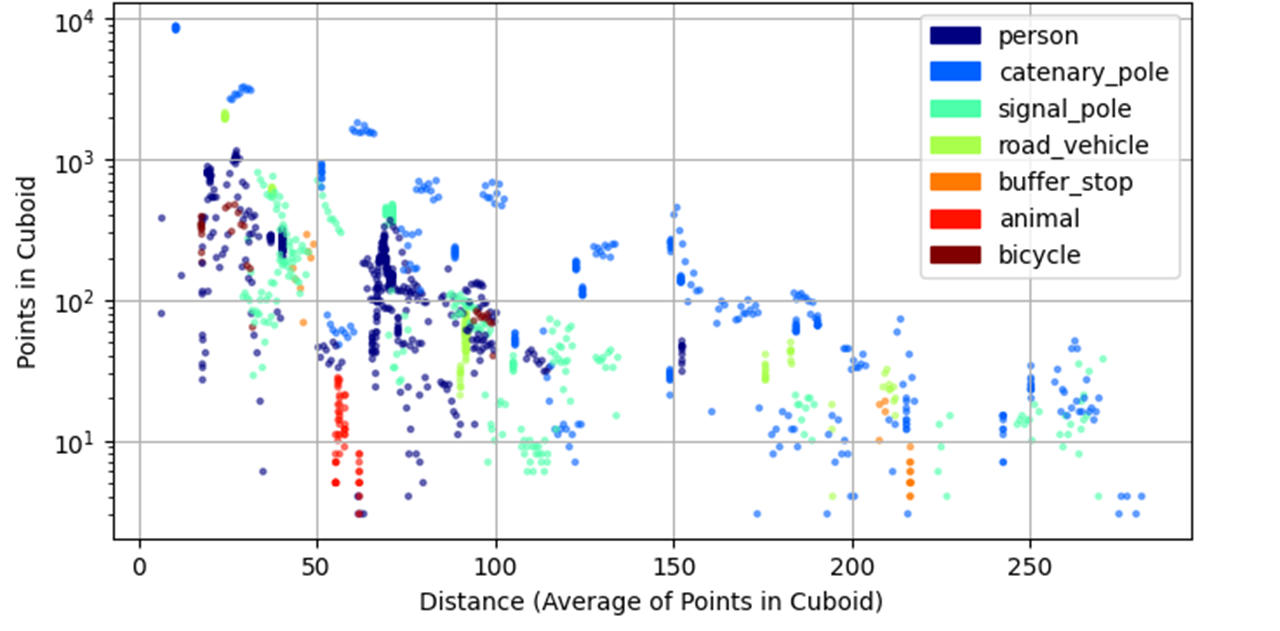}
          \end{subfigure}
        }
      \caption{Number of point cloud points within the 3D bounding box of an object, plotted against the x-axis distance from the camera to that same object. Top: Training split, Middle: Validation split, Bottom: Test split.}
      \label{fig:SplitsDists}
   \end{figure}

The detected classes include person, road vehicle, buffer stop, catenary pole, and signal pole. Fig. \ref{fig:SplitsDists} illustrates the distribution of classes in the dataset, showing the number of points inside each 3D bounding box instance in the clouds vs. the distance from the object across different splits

The distributions of the classes (number of points within the 3D bounding box of an object vs. x-axis distance from the camera to that same object) across the splits are shown in Fig. \ref{fig:SplitsDists}. Analyzing these distributions was essential for understanding the model's behavior and performance. Furthermore, the dataset labels were filtered to include only those with corresponding 2D-3D pairs, as these were the only ones relevant to this study.

\subsection{Individual Module Evaluation}

\subsubsection{2.5D Object Detection Module}

The validation results of the 2.5D object detection model are shown in Table \ref{tab:25DIndividual}. Detection quality was measured using mean average precision at an IoU threshold of 0.5 (mAP@0.5), while mean absolute error (MAE) evaluated distance prediction accuracy. The best-performing classes were person and road vehicle, likely due to their higher representation in training, particularly from KITTI pre-training. In contrast, poles were harder to detect due to their thin geometry, leading to lower mAP. MAE initially appears low within 0–250 meters (full range), but error analysis across distance ranges shows a significant increase at greater distances. This can be attributed to an imbalance in label distribution, with more short-range than long-range samples, causing the model to overfit to closer distances. Additionally, KITTI validation samples were included during pre-training, improving performance at shorter ranges.

\begin{table}[b]
\centering
\caption{Validation results of the 2.5D object detection module. P: pre-training, F: fine-tuning, K: KITTI, O: OSDaR23, A: 0-50, B: 50-100, C: 100-150, D: 150-200, E: 200-250, FR: Full Range. }
\label{tab:25DIndividual}
\renewcommand{\arraystretch}{1.2} 
\setlength{\tabcolsep}{10pt} 
\resizebox{0.485\textwidth}{!}{
\begin{tabular}{ccccccccc}
\hline
\multirow{2}{*}{\textbf{Stage}} & \multirow{2}{*}{\textbf{Dataset}} & \multirow{2}{*}{\textbf{mAP@0.5}} & \multicolumn{6}{c}{\textbf{MAE ranges (m)}} \\ 
 &  &  & \textbf{A} & \textbf{B} & \textbf{C} & \textbf{D} & \textbf{E} & \textbf{FR} \\ \hline
P & K, O & 52.0 & 9.09 & 19.37 & 23.78 & 77.41 & 136.68 & 15.09 \\
F & K & 50.0 & 9.99 & 10.08 & 18.33 & 49.21 & 75.37 & 17.10 \\ \hline
\end{tabular}%
}
\end{table}

To validate the implementation, the module was compared against Dist-YOLO \cite{app12031354} and YOLOv9, retrained using only KITTI (2D detections only). As shown in Table \ref{tab:25DKITTI}, the models exhibit comparable performance. The 2D object detection results align closely with the original implementation. However, the slightly higher MAE of our module compared to Dist-YOLO is likely due to differences in dataset splits. While YOLOv9 and our module were trained using the frequently used train/val splits mentioned in  \cite{8100174}, Dist-YOLO employed a custom split with most data allocated for training and only a few frames for validation.

\begin{table}[t]
\centering
\caption{Performance comparison between YOLOv9, Dist-YOLO, and the 2.5D object detection module from MFF on the KITTI validation dataset.}
\label{tab:25DKITTI}
\renewcommand{\arraystretch}{1.2} 
\setlength{\tabcolsep}{10pt} 
\begin{tabular}{l c c}
\hline
\textbf{Model} & \textbf{mAP@0.5} & \textbf{MAE (m)} \\ 
\hline
YOLOv9 (retrained)         & 89.9  & -     \\
Dist-YOLO (W) \cite{app12031354}  & 77.1  & 10.22 \\
MFF (Ours)     & 86.3  & 14.18 \\ 
\hline
\end{tabular}
\end{table}

\subsubsection{Depth Estimation Module}
\begin{figure}[thpb]
      \centering
      \parbox{3.3in}{
          \begin{subfigure}[b]{0.154\textwidth}
             \centering
             \includegraphics[width=\textwidth]{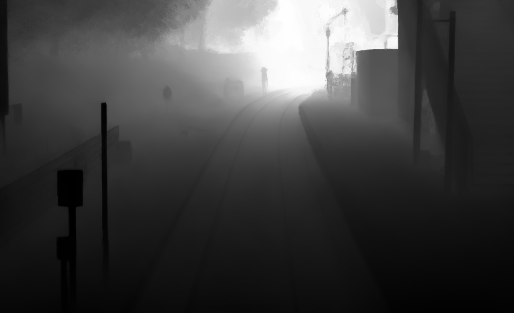}
             %\caption{}
          \end{subfigure}
          \begin{subfigure}[b]{0.154\textwidth}
             \centering
             \includegraphics[width=\textwidth]{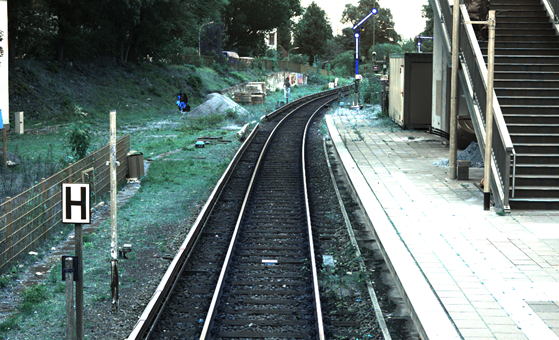}
             %\caption{}
          \end{subfigure}
          \begin{subfigure}[b]{0.154\textwidth}
             \centering
             \includegraphics[width=\textwidth]{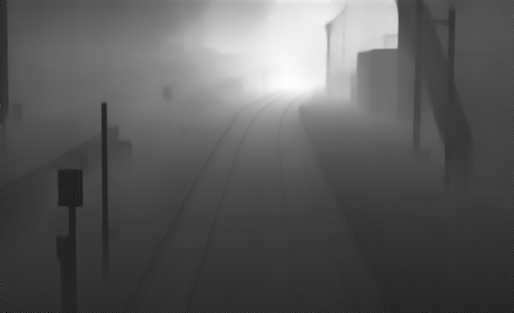}
             %\caption{}
          \end{subfigure}
        }
      \caption{Depth estimation module prediction example (relative depth). Left: Ground truth, Center: Input, and Right: Prediction.}
      \label{fig:RelDepthEstimationIndividual}
   \end{figure}

\begin{figure}[thpb]
      \centering
      \parbox{3.3in}{
          \begin{subfigure}[b]{0.154\textwidth}
             \centering
             \includegraphics[width=\textwidth]{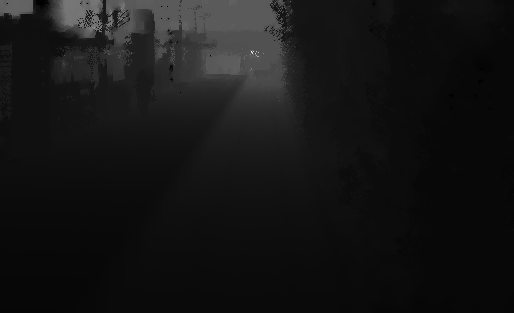}
             %\caption{}
          \end{subfigure}
          \begin{subfigure}[b]{0.154\textwidth}
             \centering
             \includegraphics[width=\textwidth]{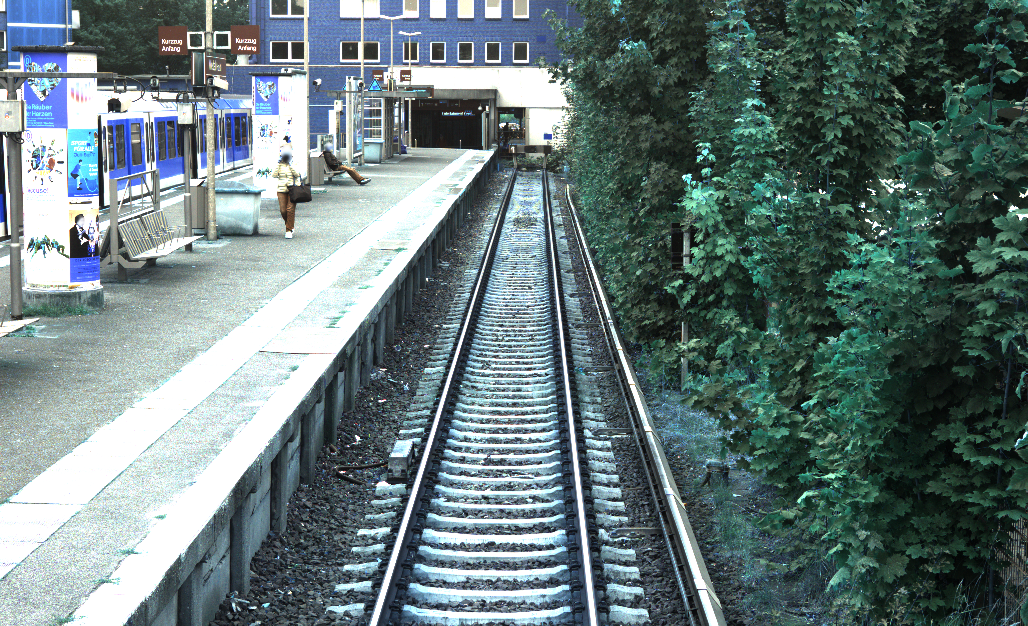}
             %\caption{}
          \end{subfigure}
          \begin{subfigure}[b]{0.154\textwidth}
             \centering
             \includegraphics[width=\textwidth]{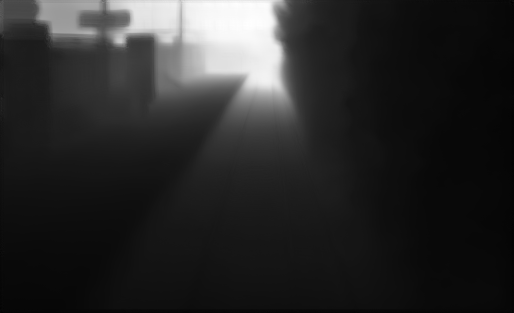}
             %\caption{}
          \end{subfigure}
        }
      \caption{Depth estimation module prediction example (absolute depth). Left: Ground truth, Center: Input, and Right: Prediction.}
      \label{fig:AbsDepthEstimationIndividual}
   \end{figure}

The depth estimation module achieved a relative depth estimation error of 0.0335, demonstrating accurate results and successful distillation (Fig. \ref{fig:RelDepthEstimationIndividual}). During fine-tuning for absolute depth estimation, a MAE of 11.51 meters was obtained. This is notable given the diverse depth range of inpainted OSDaR23 depth maps, which extended up to 500 m, requiring significant adaptation by the module (Fig. \ref{fig:AbsDepthEstimationIndividual}).

\subsubsection{Long Range Head}
To train the long-range head, only OSDaR23 data was used. Using OSDaR23-based depth maps and ground truth (GT) 2D object labels, frustums were generated, filtered by object distance, and then projected into BEV maps. As shown in Table \ref{tab:LRH}, the best-detected class was buffer stop, likely due to the limited number of examples, which were highly similar between the training and validation sets. The next best-performing classes were person and catenary pole, which had the most training samples and were well-distributed across the long-range, improving detection robustness. In contrast, signal poles, similar to the 2.5D detection case, were difficult to detect due to their small and thin structure. Notably, road vehicles in the validation set appeared only in the 200–250 m range, but the head failed to detect them. The highest MAE in long-range detection was 6.88 m, which is reasonable given detections extended up to 250 m.

\begin{table}[t]
\caption{Validation results of the long-range head. MAE is computed at different ranges. N/A: Not applicable because there where no objects of that class at that range, N/D: No detections even when there were objects of that class in a specific range.}
\label{tab:LRH}
\renewcommand{\arraystretch}{1.2} 
\setlength{\tabcolsep}{10pt} 
\resizebox{0.485\textwidth}{!}{
\centering
\begin{tabular}{cccccc}
\hline
\multirow{2}{*}{\textbf{Class}} & \multicolumn{2}{c}{\textbf{mAP@0.1}} & \multicolumn{3}{c}{\textbf{MAE (m)}} \\
 & \textbf{BEV} & \textbf{3D} & \textbf{100-150} & \textbf{150-200} & \textbf{200-250} \\ \hline
Person & \multicolumn{1}{c}{45.14} & 41.07 & \multicolumn{1}{c}{1.11} & \multicolumn{1}{c}{0.96} & 1.05 \\
Road Vehicle & \multicolumn{1}{c}{N/D} & N/D & \multicolumn{1}{c}{N/A} & \multicolumn{1}{c}{N/A} & N/D \\
Buffer Stop & \multicolumn{1}{c}{99.8} & 99.1 & \multicolumn{1}{c}{0.65} & \multicolumn{1}{c}{N/A} & N/A \\
Catenary Pole & \multicolumn{1}{c}{50.86} & 49.13 & \multicolumn{1}{c}{2.22} & \multicolumn{1}{c}{2.85} & 5.27 \\
Signal Pole & \multicolumn{1}{c}{2.79} & 1.18 & \multicolumn{1}{c}{3.91} & \multicolumn{1}{c}{5.05} & 6.88 \\ \hline
\end{tabular}
}
\end{table}

To validate the module's implementation, it was tested on the KITTI dataset (Car class) and compared to its original implementation. As shown in Table \ref{tab:LRHKITTI}, the module outperforms the original long-range head, primarily due to the increased capacity of the long-range head in MFF.

\begin{table}[b]
\centering
\caption{Performance comparison between the long-range head utilized in Faraway-Frustum and MFF. The utlized dataset was KITTI and the evaluated class was Car.}
\label{tab:LRHKITTI}
\renewcommand{\arraystretch}{1.2} 
\setlength{\tabcolsep}{10pt} 
\begin{tabular}{cc}
\hline
\textbf{Method} & \textbf{3DmAP@0.1} \\ \hline
Faraway-Frustum \cite{zhangFarawayFrustumDealingLidar2021} & 46.90     \\
MFF (ours)            & 60.61    \\ \hline
\end{tabular}%
\end{table}

\subsubsection{Short Range Head}

The short-range head was evaluated using three LiDAR-based 3D object detectors: PointRCNN, PointPillars, and Part-A$^2$. These models were trained on short-range frustums from the OSDaR23 dataset, with results shown in Table \ref{tab:SRH}. All models exhibited similar behavior. Buffer stop, road vehicle, and catenary pole achieved high detection performance due to their large, distinguishable geometry. In contrast, person and signal pole were consistently difficult to detect, likely due to noise from inpainted depth maps and the small size of these objects. Among the evaluated models, PointPillars achieved the best overall performance and was integrated into the full pipeline.

\begin{table}[b]
\centering
\caption{Validation results of different LiDAR-based 3D object detectors with OSDaR23. Based on its performance, the $\bigstar$ detector was chosen to be utilized in the full pipeline.}
\label{tab:SRH}
\renewcommand{\arraystretch}{1.2} 
\setlength{\tabcolsep}{10pt} 
\begin{tabular}{cccc}
\hline
\textbf{mAP@0.5} & \textbf{PointRCNN} & \textbf{PointPillars $\bigstar$} & \textbf{Part-A$^2$} \\ \hline
BEV & 23.624 & 31.51 & 23.89 \\
3D & 21.02 & 27.454 & 21.55 \\ \hline
\end{tabular}
\end{table}

\subsection{Full Pipeline Evaluation}

After assembling the full pipeline, it was evaluated on the OSDaR23 test set. The 2.5D object detection module achieved the results shown in Table \ref{tab:25DFull}, while the depth estimation module reached an MAE of 12.85 m. Performance closely matches validation results, which is expected given the significant overlap between training and test sequences. As previously noted, road vehicle and person, the most frequent classes, performed best, suggesting that a larger dataset could improve performance for other classes. The buffer stop MAE is the main concern, as its distribution differs between validation and test sets. In validation, buffer stops are mostly within 150 m, whereas in testing, they extend beyond 200 m, leading to higher errors.

\begin{table}[t]
\centering
\caption{Performance of the 2.5D object detection module on the OSDaR23 test set.}
\label{tab:25DFull}
\renewcommand{\arraystretch}{1.2} 
\setlength{\tabcolsep}{10pt} 
\begin{tabular}{ccc}
\hline
\textbf{Class} & \textbf{mAP@0.5} & \textbf{MAE (m)} \\ \hline
Person & 76.72 & 10.02 \\
Road Vehicle & 96.81 & 26.01 \\
Buffer Stop & 26.95 & 119.75 \\
Catenary Pole & 44.43 & 4.69 \\
Signal Pole & 58.76 & 4.77 \\
All & 60.73 & 15.76 \\ \hline
\end{tabular}
\end{table}

\begin{figure}[b]
      \centering
      \parbox{3in}{
        \includegraphics[width=0.45\textwidth]{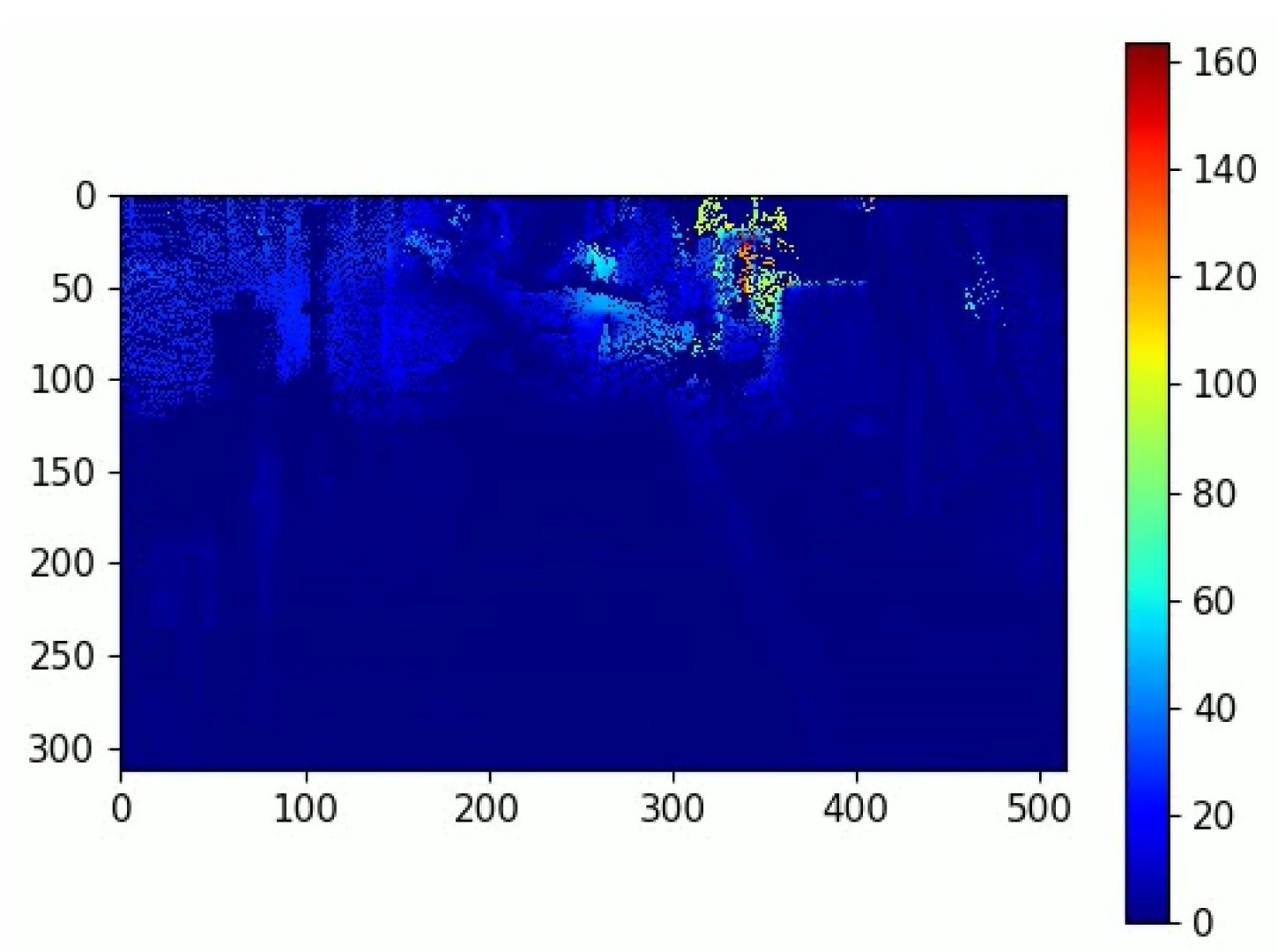}
      }
      \caption{Heatmap exemplifying the error distribution in the predicted depth maps. Error increases considerably in very distant regions.}
      \label{fig:heatmap}
   \end{figure}

Although the depth estimation module maintained similar performance, analyzing the absolute per-pixel difference between predicted and ground truth depth maps (Fig. \ref{fig:heatmap}) reveals key insights. At short range, scene structures are consistent, enabling accurate depth estimation. However, at long range, scene variations increase—some extend up to 300 m, others to 500 m—making depth estimation more challenging due to these dynamic changes.

It can be concluded that the error carried from the 2.5D predictions and depth maps can end up affecting the quality of the frustums, which at the same time ends up hindering 3D object predictions. Nevertheless, it has been shown that the pipeline works and is promising since the results obtained from individual tests demonstrate that with better distance estimations as well as less noisy and more accurate depth maps, the pipeline would be capable of producing accurate results.

\begin{figure}[t]
      \centering
      \parbox{3.3in}{
          \centering
          \begin{subfigure}[b]{0.45\textwidth}
             \centering
             \includegraphics[width=\textwidth]{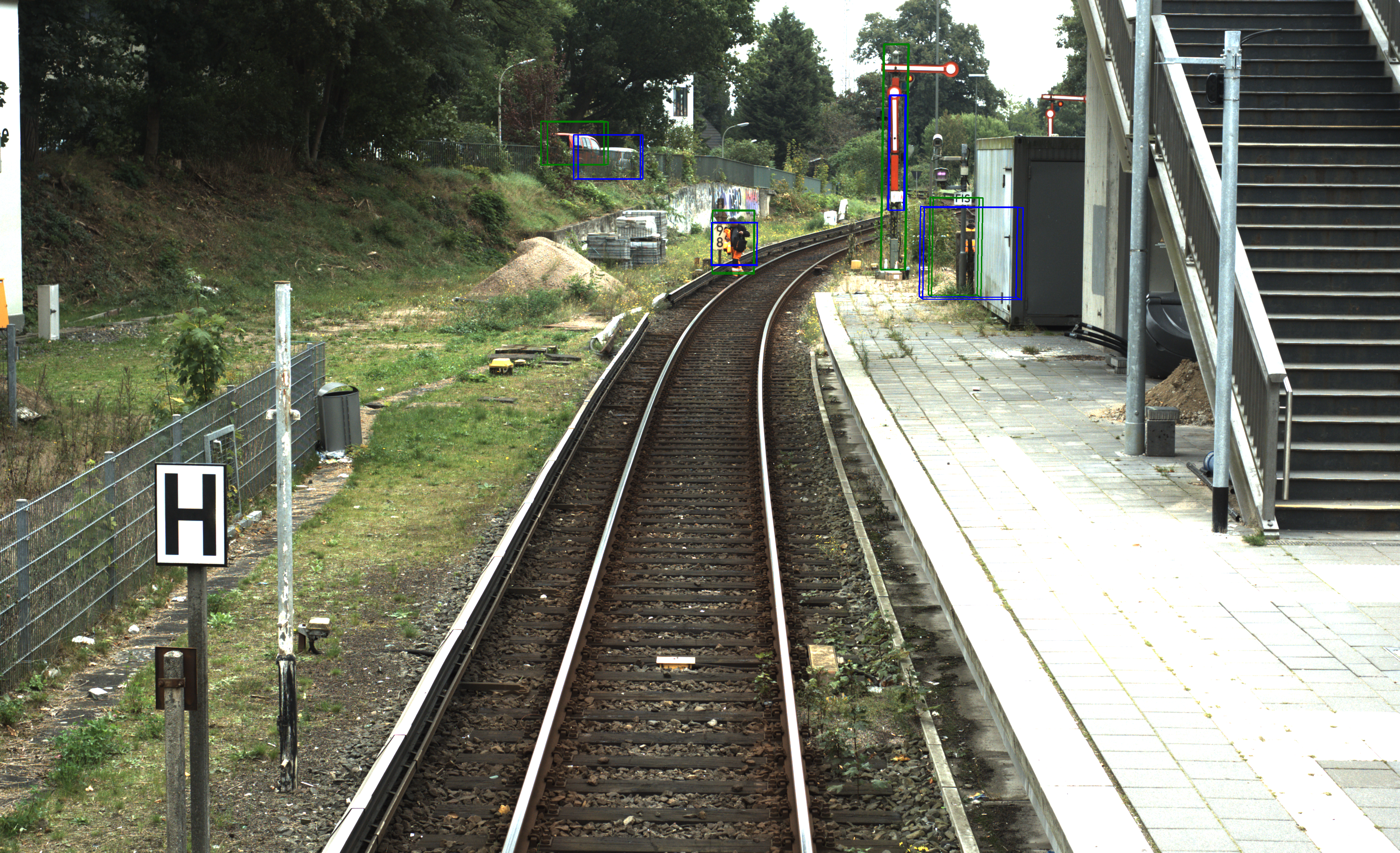}
          \end{subfigure}
          \begin{subfigure}[b]{0.45\textwidth}
             \centering
             \includegraphics[width=\textwidth]{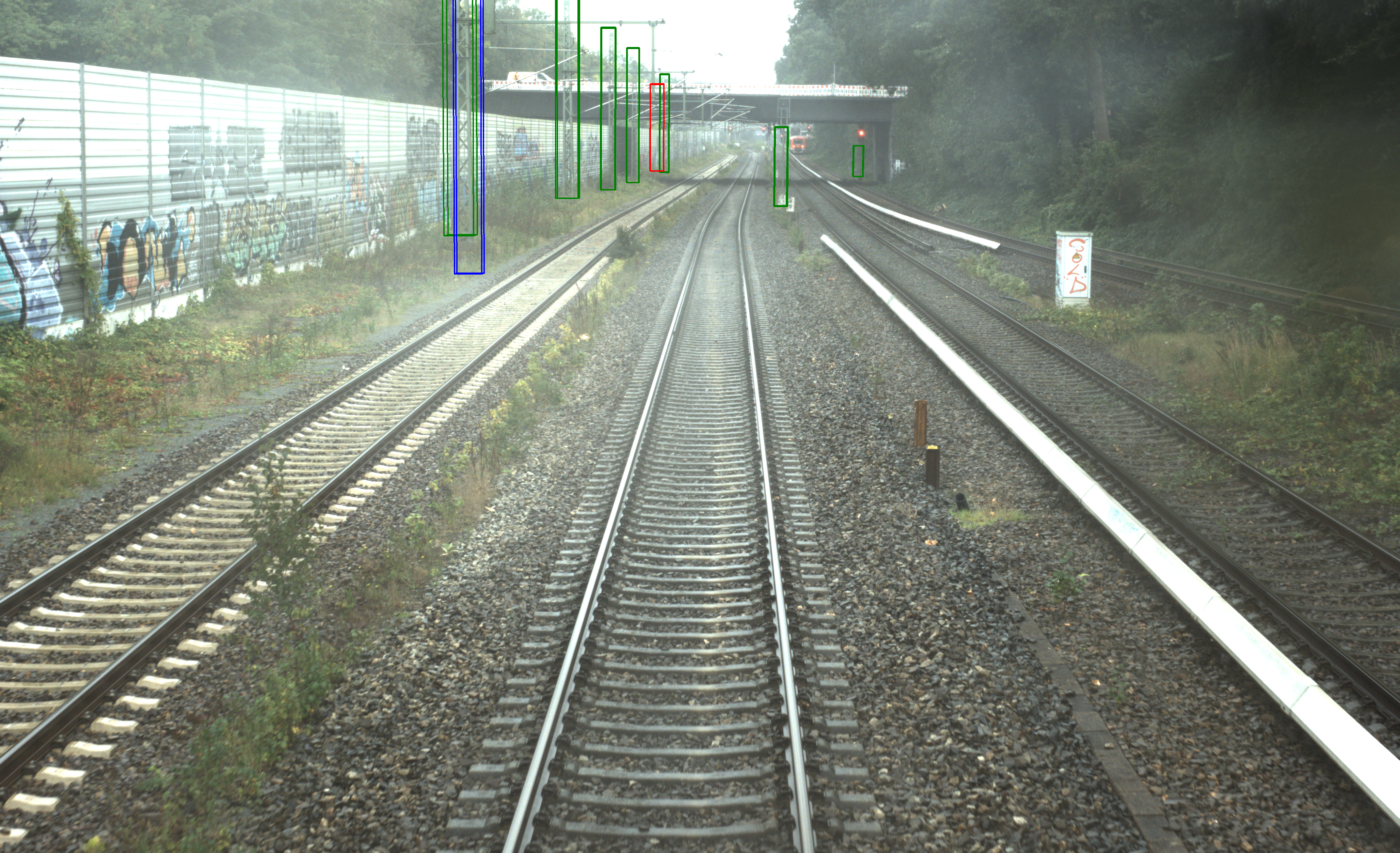}
          \end{subfigure}
        }
      \caption{Examples of detections made by MFF, green boxes represent GTs,blue boxes represent short-range 3D object predictions and red boxes represent long-range 3D object predictions.}
      \label{fig:FinalResults}
   \end{figure}

As shown in Fig. \ref{fig:FinalResults}, the full pipeline successfully produces 3D object detections. The top image demonstrates accurate detections of persons, vehicles, and a pole. However, in the bottom image, while both short- and long-range detections are present, some poles remain undetected. This occurs because the first image contains classes that the 2.5D detection module handles well, whereas the second image consists of objects the module struggles to detect.

These results indicate that errors from 2.5D predictions and depth maps degrade frustum quality, ultimately hindering 3D object predictions. However, the pipeline remains effective, and improving depth estimation and reducing noise would potentially further enhance its performance.

\section{CONCLUSIONS}
This work explored the challenges and advancements in achieving fully autonomous trains, with a focus on long-range 3D object detection. The focus on three main objectives: developing an AI-based 3D object detection system using only monocular images, enabling long-range detection, and improving accuracy through LiDAR guidance. It introduces a modular detection method combining a 2.5D object detector, depth estimation, and a decision mechanism for short- and long-range detection. Experimental results show that while the approach is effective, limitations in data distribution, depth accuracy, and backbone module performance affect precision. Future work suggests real-time performance optimization, end-to-end training, synthetic data integration, noise reduction in depth maps, and testing other 2.5D object detection models to enhance accuracy and robustness. Despite these challenges, the approach establishes a solid foundation for future research and contributes to advancing autonomous railway perception.

%\addtolength{\textheight}{-12cm}   % This command serves to balance the column lengths
                                  % on the last page of the document manually. It shortens
                                  % the textheight of the last page by a suitable amount.
                                  % This command does not take effect until the next page
                                  % so it should come on the page before the last. Make
                                  % sure that you do not shorten the textheight too much.

%%%%%%%%%%%%%%%%%%%%%%%%%%%%%%%%%%%%%%%%%%%%%%%%%%%%%%%%%%%%%%%%%%%%%%%%%%%%%%%%

%%%%%%%%%%%%%%%%%%%%%%%%%%%%%%%%%%%%%%%%%%%%%%%%%%%%%%%%%%%%%%%%%%%%%%%%%%%%%%%%

%%%%%%%%%%%%%%%%%%%%%%%%%%%%%%%%%%%%%%%%%%%%%%%%%%%%%%%%%%%%%%%%%%%%%%%%%%%%%%%%
%\section*{APPENDIX}

%Appendixes should appear before the acknowledgment.

\section*{ACKNOWLEDGMENT}

This work has received funding from the German Federal Ministry for Economic Affairs and Climate Action (BMWK) under grant agreement 19I21039A.

%%%%%%%%%%%%%%%%%%%%%%%%%%%%%%%%%%%%%%%%%%%%%%%%%%%%%%%%%%%%%%%%%%%%%%%%%%%%%%%%

\bibliographystyle{Bibliography/IEEEtran}
\bibliography{Bibliography/IEEEabrv,Bibliography/IEEEexample}

\begin{thebibliography}{10}
\providecommand{\url}[1]{#1}
\csname url@rmstyle\endcsname
\providecommand{\newblock}{\relax}
\providecommand{\bibinfo}[2]{#2}
\providecommand\BIBentrySTDinterwordspacing{\spaceskip=0pt\relax}
\providecommand\BIBentryALTinterwordstretchfactor{4}
\providecommand\BIBentryALTinterwordspacing{\spaceskip=\fontdimen2\font plus
\BIBentryALTinterwordstretchfactor\fontdimen3\font minus \fontdimen4\font\relax}
\providecommand\BIBforeignlanguage[2]{{%
\expandafter\ifx\csname l@#1\endcsname\relax
\typeout{** WARNING: IEEEtran.bst: No hyphenation pattern has been}%
\typeout{** loaded for the language `#1'. Using the pattern for}%
\typeout{** the default language instead.}%
\else
\language=\csname l@#1\endcsname
\fi
#2}}

\bibitem{Borghini2021}
C.~Borghini, ``Trains: The backbone of mobility,'' \emph{Pais Positivo}, 2021, originally published in Portuguese in the April edition of the Pais Positivo magazine, issue no 144.

\bibitem{digitale_schiene_deutschland}
``What is digitale schiene deutschland?'' \url{https://digitale-schiene-deutschland.de/en/What-is-Digitale-Schiene-Deutschland}.

\bibitem{shift2rail}
``About shift2rail,'' \url{https://rail-research.europa.eu/about-shift2rail/}.

\bibitem{safetrain}
\BIBentryALTinterwordspacing
{safe.trAIn}, ``safe.train: Rethinking mobility.'' [Online]. Available: \url{https://safetrain-projekt.de/en/}
\BIBentrySTDinterwordspacing

\bibitem{RailwaySafetyStatistics}
\BIBentryALTinterwordspacing
(2023) Railway safety statistics in the {{EU}}. [Online]. Available: \url{https://ec.europa.eu/eurostat/statistics-explained/index.php?title=Railway\_safety\_statistics\_in\_the\_EU}
\BIBentrySTDinterwordspacing

\bibitem{ristic-durrantReviewVisionBasedOnBoard2021a}
\BIBentryALTinterwordspacing
D.~Ristić-Durrant, M.~Franke, and K.~Michels, ``A {{Review}} of {{Vision-Based On-Board Obstacle Detection}} and {{Distance Estimation}} in {{Railways}},'' \emph{Sensors}, vol.~21, no.~10, p. 3452, 2021. [Online]. Available: \url{https://www.mdpi.com/1424-8220/21/10/3452}
\BIBentrySTDinterwordspacing

\bibitem{assafHighPrecisionLowCostGimballing2022}
\BIBentryALTinterwordspacing
E.~H. Assaf, C.~von Einem, C.~Cadena, R.~Siegwart, and F.~Tschopp, ``High-{{Precision Low-Cost Gimballing Platform}} for {{Long-Range Railway Obstacle Detection}},'' \emph{Sensors}, vol.~22, no.~2, p. 474, 2022. [Online]. Available: \url{https://www.mdpi.com/1424-8220/22/2/474}
\BIBentrySTDinterwordspacing

\bibitem{yang2024improvingdistant3dobject}
\BIBentryALTinterwordspacing
Z.~Yang, Z.~Yu, C.~Choy, R.~Wang, A.~Anandkumar, and J.~M. Alvarez, ``Improving distant 3d object detection using 2d box supervision,'' 2024. [Online]. Available: \url{https://arxiv.org/abs/2403.09230}
\BIBentrySTDinterwordspacing

\bibitem{Wang2023}
\BIBentryALTinterwordspacing
Y.~Wang, ``Lidar on its way out? camera's market size from 76\% to 79\% by 2033,'' \emph{IDTechEx research}, 2023. [Online]. Available: \url{https://www.idtechex.com/en/research-article/lidar-on-its-way-out-cameras-market-size-from-76-to-79-by-2033/28865}
\BIBentrySTDinterwordspacing

\bibitem{wuVirtualSparseConvolution2023}
H.~Wu, C.~Wen, S.~Shi, X.~Li, and C.~Wang, ``Virtual {{Sparse Convolution}} for {{Multimodal 3D Object Detection}},'' in \emph{{{CVPR}} 2023}, 2023, pp. 21\,653--21\,662.

\bibitem{liLoGoNetAccurate3D2023}
X.~Li, T.~Ma, Y.~Hou, B.~Shi, Y.~Yang, Y.~Liu, X.~Wu, Q.~Chen, Y.~Li, Y.~Qiao, and L.~He, ``{{LoGoNet}}: {{Towards Accurate 3D Object Detection With Local-to-Global Cross-Modal Fusion}},'' in \emph{{{CVPR}} 2023}, 2023, pp. 17\,524--17\,534.

\bibitem{liVoxelFieldFusion2022}
Y.~Li, X.~Qi, Y.~Chen, L.~Wang, Z.~Li, J.~Sun, and J.~Jia, ``Voxel {{Field Fusion}} for {{3D Object Detection}},'' in \emph{{{CVPR}} 2022}, 2022, pp. 1120--1129.

\bibitem{liHomogeneousMultimodalFeature2022}
X.~Li, B.~Shi, Y.~Hou, X.~Wu, T.~Ma, Y.~Li, and L.~He, ``Homogeneous {{Multi-modal Feature Fusion}} and~{{Interaction}} for~{{3D Object Detection}},'' in \emph{{{ECCV}} 2022}, S.~Avidan, G.~Brostow, M.~Ciss{\'e}, G.~M. Farinella, and T.~Hassner, Eds.\hskip 1em plus 0.5em minus 0.4em\relax Cham: Springer Nature Switzerland, 2022, pp. 691--707.

\bibitem{OLEKSIIENKO2022313}
I.~Oleksiienko and A.~Iosifidis, ``Chapter 13 - {{3D}} object detection and tracking,'' in \emph{Deep Learning for Robot Perception and Cognition}, A.~Iosifidis and A.~Tefas, Eds.\hskip 1em plus 0.5em minus 0.4em\relax Academic Press, 2022, pp. 313--340.

\bibitem{zhangMonoDETRDepthguidedTransformer2023}
R.~Zhang, H.~Qiu, T.~Wang, Z.~Guo, Z.~Cui, Y.~Qiao, H.~Li, and P.~Gao, ``{{MonoDETR}}: {{Depth-guided Transformer}} for {{Monocular 3D Object Detection}},'' in \emph{{{ICCV}} 2023}, 2023, pp. 9155--9166.

\bibitem{youPseudoLiDARAccurateDepth2020}
Y.~You, Y.~Wang, W.-L. Chao, D.~Garg, G.~Pleiss, B.~Hariharan, M.~Campbell, and K.~Q. Weinberger, ``Pseudo-{{LiDAR}}++: {{Accurate Depth}} for {{3D Object Detection}} in {{Autonomous Driving}},'' Feb. 2020.

\bibitem{liBEVDepthAcquisitionReliable2023}
Y.~Li, Z.~Ge, G.~Yu, J.~Yang, Z.~Wang, Y.~Shi, J.~Sun, and Z.~Li, ``{{BEVDepth}}: {{Acquisition}} of {{Reliable Depth}} for {{Multi-View 3D Object Detection}},'' \emph{Proceedings of the AAAI Conference on Artificial Intelligence}, vol.~37, no.~2, pp. 1477--1485, June 2023.

\bibitem{li2024unified3dobjectdetection}
\BIBentryALTinterwordspacing
Z.~Li, X.~Xu, S.~Lim, and H.~Zhao, ``Towards unified 3d object detection via algorithm and data unification,'' 2024. [Online]. Available: \url{https://arxiv.org/abs/2402.18573}
\BIBentrySTDinterwordspacing

\bibitem{RAULPAPER}
M.~Mahmeen, R.~D. Dominguez~Sanchez, M.~Friebe, M.~Pech, and S.~Haider, ``Collision avoidance route planning for autonomous medical devices using multiple depth cameras,'' \emph{IEEE Access}, vol.~10, pp. 29\,903--29\,915, 2022.

\bibitem{qiFrustumPointNets3D2018}
C.~R. Qi, W.~Liu, C.~Wu, H.~Su, and L.~J. Guibas, ``Frustum {{PointNets}} for {{3D Object Detection From RGB-D Data}},'' in \emph{{{CVPR}} 2018}, 2018, pp. 918--927.

\bibitem{zhangFarawayFrustumDealingLidar2021}
H.~Zhang, D.~Yang, E.~Yurtsever, K.~A. Redmill, and {\"U}.~{\"O}zg{\"u}ner, ``Faraway-{{Frustum}}: {{Dealing}} with {{Lidar Sparsity}} for {{3D Object Detection}} using {{Fusion}},'' Mar. 2021.

\bibitem{shi2021pvrcnnpointvoxelfeatureset}
\BIBentryALTinterwordspacing
S.~Shi, C.~Guo, L.~Jiang, Z.~Wang, J.~Shi, X.~Wang, and H.~Li, ``Pv-rcnn: Point-voxel feature set abstraction for 3d object detection,'' 2021. [Online]. Available: \url{https://arxiv.org/abs/1912.13192}
\BIBentrySTDinterwordspacing

\bibitem{howard2017mobilenetsefficientconvolutionalneural}
\BIBentryALTinterwordspacing
A.~G. Howard, M.~Zhu, B.~Chen, D.~Kalenichenko, W.~Wang, T.~Weyand, M.~Andreetto, and H.~Adam, ``Mobilenets: Efficient convolutional neural networks for mobile vision applications,'' 2017. [Online]. Available: \url{https://arxiv.org/abs/1704.04861}
\BIBentrySTDinterwordspacing

\bibitem{KITTI}
A.~Geiger, P.~Lenz, and R.~Urtasun, ``Are we ready for autonomous driving? the kitti vision benchmark suite,'' in \emph{Conference on Computer Vision and Pattern Recognition (CVPR)}, 2012.

\bibitem{7780460}
J.~Redmon, S.~Divvala, R.~Girshick, and A.~Farhadi, ``You only look once: Unified, real-time object detection,'' in \emph{2016 IEEE Conference on Computer Vision and Pattern Recognition (CVPR)}, 2016, pp. 779--788.

\bibitem{wang2024yolov9learningwantlearn}
\BIBentryALTinterwordspacing
C.-Y. Wang, I.-H. Yeh, and H.-Y.~M. Liao, ``Yolov9: Learning what you want to learn using programmable gradient information,'' 2024. [Online]. Available: \url{https://arxiv.org/abs/2402.13616}
\BIBentrySTDinterwordspacing

\bibitem{wang2024yolov1yolov10fastestaccurate}
\BIBentryALTinterwordspacing
C.-Y. Wang and H.-Y.~M. Liao, ``Yolov1 to yolov10: The fastest and most accurate real-time object detection systems,'' 2024. [Online]. Available: \url{https://arxiv.org/abs/2408.09332}
\BIBentrySTDinterwordspacing

\bibitem{haseebMachineLearningTechniques2021}
M.~A. Haseeb, ``Machine learning techniques for autonomous multi-sensor long-range environmental perception system,'' Ph.D. dissertation, Universität Bremen, 2021.

\bibitem{app12031354}
\BIBentryALTinterwordspacing
M.~Vajgl, P.~Hurtik, and T.~Nejezchleba, ``Dist-yolo: Fast object detection with distance estimation,'' \emph{Applied Sciences}, vol.~12, no.~3, 2022. [Online]. Available: \url{https://www.mdpi.com/2076-3417/12/3/1354}
\BIBentrySTDinterwordspacing

\bibitem{Masoumian_2021}
\BIBentryALTinterwordspacing
A.~Masoumian, D.~G. Marei, S.~Abdulwahab, J.~Cristiano, D.~Puig, and H.~A. Rashwan, \emph{Absolute Distance Prediction Based on Deep Learning Object Detection and Monocular Depth Estimation Models}.\hskip 1em plus 0.5em minus 0.4em\relax IOS Press, Oct. 2021. [Online]. Available: \url{http://dx.doi.org/10.3233/FAIA210151}
\BIBentrySTDinterwordspacing

\bibitem{ranftl2020robustmonoculardepthestimation}
\BIBentryALTinterwordspacing
R.~Ranftl, K.~Lasinger, D.~Hafner, K.~Schindler, and V.~Koltun, ``Towards robust monocular depth estimation: Mixing datasets for zero-shot cross-dataset transfer,'' 2020. [Online]. Available: \url{https://arxiv.org/abs/1907.01341}
\BIBentrySTDinterwordspacing

\bibitem{Alhashim2018}
\BIBentryALTinterwordspacing
I.~Alhashim and P.~Wonka, ``High quality monocular depth estimation via transfer learning,'' \emph{arXiv e-prints}, vol. abs/1812.11941, 2018. [Online]. Available: \url{https://arxiv.org/abs/1812.11941}
\BIBentrySTDinterwordspacing

\bibitem{huang2018denselyconnectedconvolutionalnetworks}
\BIBentryALTinterwordspacing
G.~Huang, Z.~Liu, L.~van~der Maaten, and K.~Q. Weinberger, ``Densely connected convolutional networks,'' 2018. [Online]. Available: \url{https://arxiv.org/abs/1608.06993}
\BIBentrySTDinterwordspacing

\bibitem{levin2004colorization}
A.~Levin, D.~Lischinski, and Y.~Weiss, ``Colorization using optimization,'' in \emph{ACM SIGGRAPH 2004 Papers}, 2004, pp. 689--694.

\bibitem{ke2023repurposing}
B.~Ke, A.~Obukhov, S.~Huang, N.~Metzger, R.~C. Daudt, and K.~Schindler, ``Repurposing diffusion-based image generators for monocular depth estimation,'' in \emph{Proceedings of the IEEE/CVF Conference on Computer Vision and Pattern Recognition (CVPR)}, 2024.

\bibitem{nuscenes}
H.~Caesar, V.~Bankiti, A.~H. Lang, S.~Vora, V.~E. Liong, Q.~Xu, A.~Krishnan, Y.~Pan, G.~Baldan, and O.~Beijbom, ``nuscenes: A multimodal dataset for autonomous driving,'' in \emph{CVPR}, 2020.

\bibitem{WOD}
\BIBentryALTinterwordspacing
J.~Mei, A.~Z. Zhu, X.~Yan, H.~Yan, S.~Qiao, Y.~Zhu, L.-C. Chen, H.~Kretzschmar, and D.~Anguelov, ``Waymo open dataset: Panoramic video panoptic segmentation,'' 2022. [Online]. Available: \url{https://arxiv.org/abs/2206.07704}
\BIBentrySTDinterwordspacing

\bibitem{zendelRailSem19DatasetSemantic2019}
O.~Zendel, M.~Murschitz, M.~Zeilinger, D.~Steininger, S.~Abbasi, and C.~Beleznai, ``{{RailSem19}}: {{A Dataset}} for {{Semantic Rail Scene Understanding}},'' in \emph{Proceedings of the {{IEEE}}/{{CVF Conference}} on {{Computer Vision}} and {{Pattern Recognition Workshops}}}, 2019.

\bibitem{8859360}
Y.~Wang, L.~Wang, Y.~H. Hu, and J.~Qiu, ``Railnet: A segmentation network for railroad detection,'' \emph{IEEE Access}, vol.~7, pp. 143\,772--143\,779, 2019.

\bibitem{harb2020frsignlargescaletrafficlight}
\BIBentryALTinterwordspacing
J.~Harb, N.~Rébéna, R.~Chosidow, G.~Roblin, R.~Potarusov, and H.~Hajri, ``Frsign: A large-scale traffic light dataset for autonomous trains,'' 2020. [Online]. Available: \url{https://arxiv.org/abs/2002.05665}
\BIBentrySTDinterwordspacing

\bibitem{9715050}
L.~Guan, L.~Jia, Z.~Xie, and C.~Yin, ``A lightweight framework for obstacle detection in the railway image based on fast region proposal and improved yolo-tiny network,'' \emph{IEEE Transactions on Instrumentation and Measurement}, vol.~71, pp. 1--16, 2022.

\bibitem{10052883}
A.~Zouaoui, A.~Mahtani, M.~A. Hadded, S.~Ambellouis, J.~Boonaert, and H.~Wannous, ``Railset: A unique dataset for railway anomaly detection,'' in \emph{2022 IEEE 5th International Conference on Image Processing Applications and Systems (IPAS)}, vol. Five, 2022, pp. 1--6.

\bibitem{9852821}
S.~Liu, C.~Li, T.~Yuwen, Z.~Wan, and Y.~Luo, ``A lightweight lidar-camera sensing method of obstacles detection and classification for autonomous rail rapid transit,'' \emph{IEEE Transactions on Intelligent Transportation Systems}, vol.~23, no.~12, pp. 23\,043--23\,058, 2022.

\bibitem{khemmarRoadRailwaySmart2022}
R.~Khemmar, A.~Mauri, C.~Dulompont, J.~Gajula, V.~Vauchey, M.~Haddad, and R.~Boutteau, ``Road and {{Railway Smart Mobility}}: {{A High-Definition Ground Truth Hybrid Dataset}},'' \emph{Sensors}, vol.~22, no.~10, p. 3922, Jan. 2022.

\bibitem{Neri_2022_EUVIP}
M.~Neri and F.~Battisti, ``{3D Object Detection on Synthetic Point Clouds for Railway Applications},'' in \emph{2022 10th European Workshop on Visual Information Processing (EUVIP)}, 2022, pp. 1--6.

\bibitem{osdar23}
\BIBentryALTinterwordspacing
R.~Tilly, P.~Neumaier, K.~Schwalbe, P.~Klasek, R.~Tagiew, P.~Denzler, T.~Klockau, M.~Boekhoff, and M.~Köppel, ``\BIBforeignlanguage{de}{Open sensor data for rail 2023},'' 2023. [Online]. Available: \url{https://data.fid-move.de/dataset/3d7e7406-639f-49f6-bbca-caac511b4032}
\BIBentrySTDinterwordspacing

\bibitem{Openlabel}
\BIBentryALTinterwordspacing
{Advanced Data Controls Corp., Annotell AB, Ansys Inc., Deepen AI, Deutsches Zentrum für Luft- und Raumfahrt e. V., Five, iASYS Technology Solutions Pvt. Ltd., LiangDao GmbH, Peak Solution GmbH, SAIC Motor Corporation Ltd., Tata Consultancy Services Pvt. Ltd, understandAI GmbH, Vicomtech, WMG University of Warwick}, ``\BIBforeignlanguage{en}{Asam openlabel},'' Associationation for Standardization of Automation and Measuring Systems, Standard 1.0.0, 2021. [Online]. Available: \url{https://www.asam.net/standards/detail/openlabel/}
\BIBentrySTDinterwordspacing

\bibitem{8100174}
X.~Chen, H.~Ma, J.~Wan, B.~Li, and T.~Xia, ``Multi-view 3d object detection network for autonomous driving,'' in \emph{2017 IEEE Conference on Computer Vision and Pattern Recognition (CVPR)}, 2017, pp. 6526--6534.

\end{thebibliography}

\end{document}